\documentclass[letterpaper]{article} 
\usepackage[preprint]{aaai2027}  
\usepackage[hyphens]{url}  
\usepackage{graphicx} 
\usepackage{natbib}  
\usepackage{caption} 
\usepackage{amsmath}
\usepackage{amssymb}
\usepackage{algorithm}
\usepackage{algorithmic}
\usepackage{newfloat}
\usepackage{listings}
\DeclareCaptionStyle{ruled}{labelfont=normalfont,labelsep=colon,strut=off} 
\floatstyle{ruled}
\newfloat{listing}{tb}{lst}{}
\floatname{listing}{Listing}

\usepackage{dashrule}
\newcommand{\mydashline}[2]{%
  \noalign{\vskip 0.2ex}
  \multicolumn{8}{@{}l@{}}{%
    \rlap{\hspace*{#1}\hdashrule[0.5ex]{#2}{0.4pt}{2mm 1.2mm}}%
    \rule{0pt}{1.2ex}%
  }\\[-0.8ex]
}

\usepackage{booktabs}
\usepackage{multirow}
\usepackage{array}
\usepackage[table]{xcolor}
\usepackage{bm}
\usepackage{pifont}

\usepackage{dsfont}

\definecolor{resultcolor}{RGB}{236,242,250}

\title{Evidence-Driven Dynamic Visual Selector for Efficient Long Video Understanding}
\author{
    Bo Zhang$^1$\equalcontrib,
    Wenxin Wang$^1$\equalcontrib,
    Feng Chen$^2$\corresponding,
    Zhihao Zhang$^1$,
    Zixuan Wang$^1$,\\
    Changsheng Li$^3$,
    Yinjie Lei$^{1\ddagger}$
}
\affiliations{
    \textsuperscript{\rm 1}Sichuan University, \textsuperscript{\rm 2}Adelaide University, 
    \textsuperscript{\rm 3}Beijing Institute of Technology \\

}

\begin{document}

\maketitle

\begin{abstract}Recent advancements in MLLM-based long-form video understanding have mitigated inference-time computational cost and limited context lengths by selecting query-relevant frames.
However, existing approaches predominantly rely on external proxy scorers and rigid heuristic rules, inevitably suffering from misalignment with the target MLLM’s intrinsic evidence and failing to accommodate the non-uniform spatiotemporal information density.
In this paper, we propose a fine-grained dynamic visual selection framework named \textbf{EviSelect}, grounded in the target MLLM's internal attention evidence. Our method efficiently probes visual evidence via sparse prefilling as a structured prior to guide distribution-aware dynamic sampling. 
Specifically, we efficiently approximate attention maps of the target MLLM using highly compressed visual inputs and sparse attention, well-aligned to the full counterpart.
Conditioned on three complementary attention components derived from this prior, we design a lightweight selector that not only precisely locates query-relevant timestamps but also adaptively adjusts the local sampling rate and spatial resolution.
To enable evidence-conditioned spatiotemporal sampling, we formulate the selector as a stochastic policy and optimize it via GRPO under a joint accuracy--efficiency reward. 
By rewarding correct predictions under lower visual cost through group-relative comparisons, our method encourages the policy to allocate computation dynamically according to the information density of each video.
Across three long video understanding benchmarks, EviSelect achieves superior performance compared to existing methods while reducing selected visual tokens by about 50\% and achieving a 3.9$\times$ end-to-end speedup.
\end{abstract}

\begin{links}
    \link{Project Page}{https://zhangbo135.github.io/EviSelect/}
\end{links}


\begin{figure*}[tb]
\centering
\includegraphics[width=0.82\linewidth]{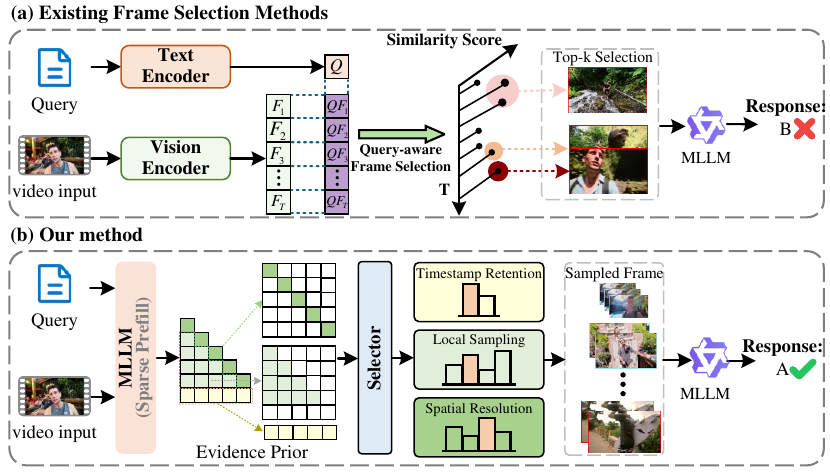} 
\caption{\textbf{Comparison of representative frame selection methods with EviSelect}. Existing method \cite{liu2025bolt, zhang2025q} typically rely on CLIP to compute query--frame similarity and apply rigid heuristic rules to select keyframes for MLLM. EviSelect rapidly approximates the target MLLM's attention maps as evidence priors and designs a selector to predict three policy distributions for each timestamp: retention decision, local sampling rate, and spatial resolution, yielding multi-scale visual input for answer generation.} 
\label{fig:abstract}
\end{figure*}

\section{Introduction}
\label{sec:intro}
Long-form video understanding \cite{cheng2025scaling, qin2025video, chen2025scaling} focuses on interpreting events over extended time scales and modeling long-range temporal dependencies in complex video content, enabling critical applications ranging from movie analysis \cite{rawal2405cinepile} to video surveillance~\cite{elharrouss2021review}. Although Multimodal Large Language Models (MLLMs) \cite{bai2025qwen2, hurst2024gpt} have demonstrated impressive performance in understanding short video clips, extending their capabilities to long-form videos faces a bottleneck. Due to the limited context windows and attention dilution \cite{qin2022devil} in long contexts, MLLMs frequently overlook query-relevant evidence and waste substantial computation on redundant frames. 

To mitigate this, recent studies \cite{liu2025bolt,yang2025hfs}, as illustrated in  Fig.~\ref{fig:abstract}(a), typically employ heuristic frame selection strategies driven by external proxies to extract query-conditioned frames. 
For example, Q-Frame \cite{zhang2025q} utilizes CLIP \cite{radford2021learning} to score query-frame similarity, performs ranking-based frame selection, and feeds the selected frames into an MLLM with a fixed resolution strategy.
However, these paradigms suffer from three critical limitations:
(1) \textit{Objective Misalignment}. Because proxy signals merely reflect semantic relevance rather than align with the target MLLM's internal evidence requirements \cite{li2025less}, existing paradigms \cite{liang2024keyvideollm,hu2025m} inevitably miss subtle but critical visual cues.
(2) \textit{Static Frame Retrieval}. Existing methods \cite{tang2025adaptive,liu2025bolt} often treat long-video understanding as an isolated keyframe selection problem, rather than modeling a query-conditioned belief distribution over continuous temporal trajectories. Consequently, these approaches lead to temporally myopic selections that over-concentration on visually salient segments while insufficient coverage of sparse yet critical events.
(3) \textit{Rigid Spatiotemporal Configuration}. Existing methods \cite{sheng2025sevices,yao2025k} adopt a one-size-fits-all sampling setup—selecting a predetermined number of frames at a fixed spatial resolution—regardless of the underlying non-uniform information density of the video. Such rigid configurations result in inefficient allocation of the computational budget, either over-sampling redundant segments or under-sampling sparse critical events.

To alleviate these challenges, we propose \textbf{EviSelect}, a fine-grained dynamic visual selection framework shown in Fig.~\ref{fig:abstract} (b).
Rather than relying on external proxies, EviSelect directly extracts endogenous evidence from attention maps, which inherently reflect its query-conditioned spatiotemporal dependencies and actual visual demands.
Specifically, we first employ a sparse pre-filling mechanism over highly compressed anchor frames to efficiently probe the target MLLM's internal attention maps at low cost.
We further decompose these maps into query--frame relevance, cross-frame temporal dependency, and intra-frame spatial evidence.
A lightweight selector then leverages the three cues to drive a joint distribution-aware sampling policy by predicting three corresponding fine-grained dimensions at each timestamp:  retention decision, local sampling rate, and spatial resolution.
To align this sampling policy with the target MLLM's actual performance, we formulate the selector as a stochastic policy and optimize it via GRPO~\cite{shao2024deepseekmath} using a joint accuracy-efficiency reward. By rewarding correct MLLM predictions achieved at lower visual cost, our method drives the policy to adaptively allocate computation according to the video's intrinsic spatiotemporal information density, rather than relying on fixed heuristics.

Our main contributions can be summarized as follows:

\begin{itemize}

\item 
We propose {EviSelect}, a fine-grained dynamic visual selection framework that reformulates keyframe selection as a distribution-aware dynamic sampling driven by internal evidence of the target MLLM.

\item 
We propose a sparse prefill mechanism to rapidly approximate the evidence prior and a selector for fine-grained, distribution-aware spatiotemporal sampling policy.

\item 
Extensive experiments on three benchmarks demonstrate that our method achieves state-of-the-art performance with only 50\% of the selected visual tokens and accelerates the end-to-end process by 3.9$\times$.

\end{itemize}

\section{Related Work}
\label{sec:related_work}

\subsection{Long-form Video Understanding.}
Existing MLLMs~\cite{bai2023qwen, hurst2024gpt} have achieved significant progress in vision-language understanding, with representative early efforts including BLIP-2~\cite{li2023blip} and Flamingo~\cite{alayrac2022flamingo}. However, long-form video understanding remains challenging for MLLM due to their limited context window and prohibitive computational cost. To this end, recent studies typically follow two paradigms: (1) token compression~\cite{sun2025llava,chen2025sparsity}, motivated by the substantial visual redundancy across video frames, introduces trainable compression modules and fine-tunes the model to aggregate visual content into fewer tokens~\cite{li2024llama, shu2025video,chen2026omnisparse, he2025zipvl}. VideoChat-Flash~\cite{li2024videochat} hierarchically compresses long videos from the clip level to the video level. (2) context extension~\cite{shen2025long}, which enables MLLMs to handle longer inputs via length extrapolation and additional training on longer videos~\cite{chen2024longvila, zhang2024long}. Long-VITA~\cite{chen2024longvila} adopts context-parallel distributed inference and a logits-masked language modeling head to scale to extremely long token sequences. However, they typically rely on additional model modifications and may struggle to preserve key information in the video.

\subsection{Query-based Keyframe Selection.}
Instead of modifying the MLLM via compression or context extension, a complementary research direction for long-form video understanding focuses on selecting query-relevant keyframes to capture critical visual information, ranging from training-free frame matching~\cite{zhu2025focus, zhang2025q} to training-based selectors~\cite{zhang2025flexselect, buch2025flexible}. Training-free methods~\cite{tang2025adaptive, liu2025bolt} typically rely on external proxies, such as CLIP~\cite{radford2021learning} or DINOv2~\cite{oquab2023dinov2}, to select informative frames. For example, DIG~\cite{li2025divide} adapts its sampling strategy based on query type and uses DINOv2 to compute pairwise frame similarity to derive candidate segments, and then leverages an MLLM-based reward signal to expand local segments. 
AKS~\cite{tang2025adaptive} uses CLIP to compute frame–query relevance scores and selects keyframes by jointly optimizing relevance and temporal coverage. 
Meanwhile, training-based selectors~\cite{hu2025m, yao2025k} model frame selection as a learnable policy. Frame-Voyager~\cite{yu2024frame} constructs subset-level supervision to query frame subsets combinatorially, rather than selecting frames independently. TSPO~\cite{tang2025tspo} proposes a trainable event-aware temporal agent that models event-query correlation for probabilistic keyframe selection. 
Despite their effectiveness, they rely on external proxy scorers and rigid heuristic rules, which suffer from misalignment with the intrinsic preferences of the target MLLM, and often fail to accommodate the dynamic spatiotemporal information density.

\subsection{Reinforcement Learning for MLLMs.}
Recent reinforcement learning methods~\cite{zhang2024direct, kim2026mc} have advanced beyond classic RLHF pipelines~\cite{ rafailov2023direct} by adopting group-based policy optimization objectives for more stable MLLM optimization. In particular, Group Relative Policy Optimization (GRPO)~\cite{shao2024deepseekmath} estimates advantages through within-group relative ranking among multiple sampled candidates for the same input within each update, reducing reliance on critic training and improving robustness under high-variance rewards. Such group-based RL formulations have been explored in a range of applications, especially for tasks emphasizing strong reasoning~\cite{wu2025spatial} or structured decision-making~\cite{feng2025group, wang2026policytrim}, where relative preference signals are easier to obtain and more stable than absolute scalar rewards. In this paper, our EviSelect takes a novel perspective that directly leverages GRPO to optimize dynamic sampling strategy that balances accuracy and token efficiency, which directly addresses the most critical bottleneck in long video understanding: extracting decisive evidence from extremely long contexts.

\section{Methodology}
\label{sec:method}

\begin{figure*}[tb]
\centering
\includegraphics[width=1\linewidth]{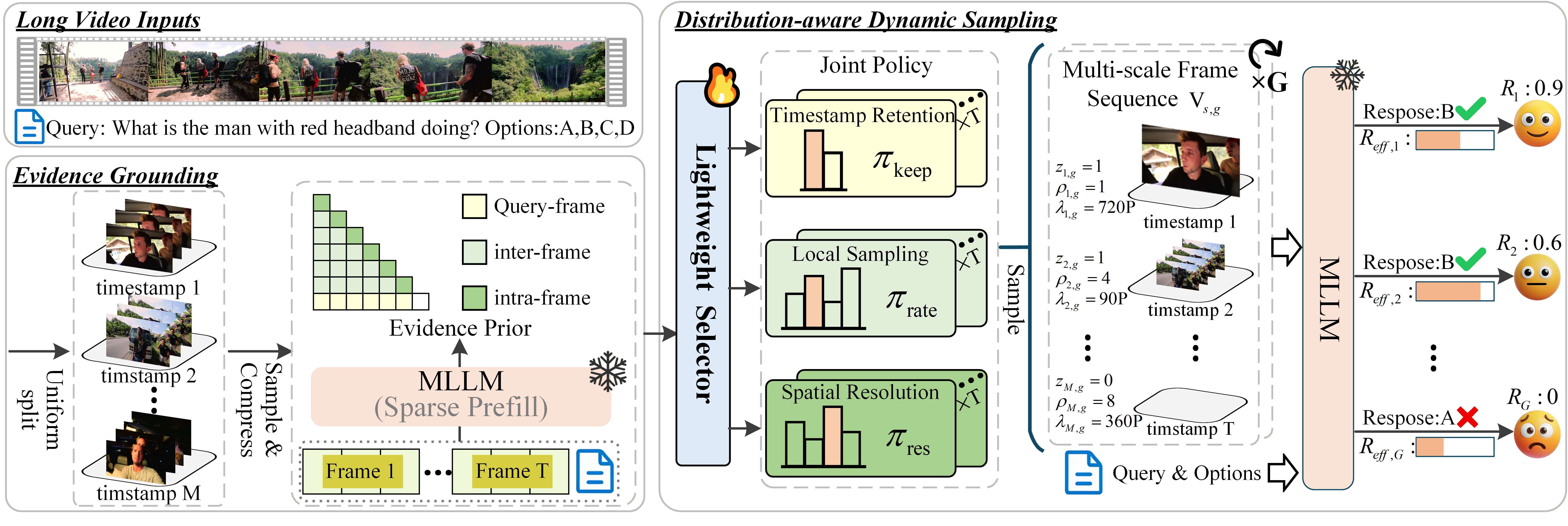} 
\caption{\textbf{Overview of our method.} 
Our method efficiently probes visual evidence to enable distribution-aware dynamic sampling. We partition the video into uniform timestamps, construct a highly compressed visual input, and employ sparse attention to rapidly approximate the MLLM's attention maps. A lightweight selector then predicts three policy distributions for each timestamp, including retention decision, sampling rate, and spatial resolution. These decisions are used to resample the video for final inference. The policy is optimized via GRPO with a joint accuracy–efficiency reward.} 
\label{fig:main}
\end{figure*}

\subsection{Overview}
As shown in Fig.~\ref{fig:main}, given a textual query $Q$ and a long video $V$, we aim to construct a compact yet informative multi-scale visual input $V_s$ that aligns with the target MLLM’s intrinsic evidence demands and adapts to the video's dynamic information density. 
We cast this process as a distribution-aware spatiotemporal sampling policy that outputs fine-grained visual allocation decisions $\mathbf{u}_t=(z_t,\rho_t,\lambda_t)$ at each timestamp $t$, representing timestamp retention, local temporal sampling rate, and spatial resolution, respectively.
Our framework efficiently captures the target MLLM's attention priors and models the joint sampling policy.
In the evidence grounding stage, we perform an efficient probing pass through the sparse prefilling of the target MLLM over a highly compressed visual token of $V$, obtaining attention maps $\mathbf{A}$ as structured priors. 
Then we decompose $\mathbf{A}$ into three complementary cues: query-frame relevance $\mathbf{A}_{qf}$ for $z_t$, cross-frame temporal dependency $\mathbf{A}_{ff}$ for $\rho_t$, and intra-frame spatial saliency $\mathbf{A}_{if}$ for $\lambda_t$. 
Conditioned on these cues, a lightweight selector directly predicts the joint distribution over $\mathbf{u}_t$, and is trained through GRPO with a joint accuracy-efficiency reward. As a result, the selector dynamically adjusts the fine-grained decisions, extracting only the essential visual context. Finally, we resample $V$ according to these decisions to produce $V_s$ for final inference by the target MLLM.

\subsection{Evidence Grounding}
We extract layer-wise attention priors from the target MLLM via an efficient probing pass on a highly compressed video.
Specifically, we perform sparse prefilling to obtain block-level sparse attentions $\{A^{(l)}\}_{l=1}^{L}$ for all $L$ layers, which are decomposed into three evidence cues $\mathbf{A}_{qf}$, $\mathbf{A}_{ff}$, and $\mathbf{A}_{if}$.

\textbf{Sparse Evidence Probing.}
Although all sampling decisions $(z_t,\rho_t,\lambda_t)$ are operated at the frame level, they require evidence at different granularities: temporal retention and local temporal sampling mainly depend on frame-level query relevance and cross-frame dependency, whereas spatial-resolution selection requires within-frame token structure.
We therefore obtain these cues through an efficient sparse prefilling pass~\cite{fan2026flashprefill,lai2025flexprefill}, circumventing the heavy computational cost of reconstructing dense token-level attention over the full video.

We first uniformly partition the video timeline into consecutive temporal segments and use the midpoint of each segment as an anchor timestamp, yielding $T$ anchors. Each anchor frame is then spatially downsampled, and the resulting frames form the compressed visual input $V_c$.
The target MLLM encodes $V_c$ into visual tokens, which are concatenated with the query tokens of $Q$ to form the prefilling context sequence $X$.
During prefilling stage, we employ block-sparse attention~\cite{lu2025moba} to efficiently obtain the attention maps and apply adaptive top-$p$ retention~\cite{he2024zipvl}, which ensures that the retained key blocks cover at least $p$ of the attention mass. By constraining the attention mass discarded, top-$p$ retention strategy reduces the perturbation introduced by sparsification~\cite{lin2025twilight}, so the resulting attention maps closely approximate those of full attention. We further validate this similarity in Tab.~\ref{tab:sparse_and_dense}.
Formally, this probing process is abstracted as: 

\begin{equation}
\left\{\mathbf{A}^{(l)}\right\}
= \operatorname{SparsePrefill}(X;\tau_p,B),
\end{equation}
where $l\in\{1,\ldots,L\}$ index the transformer layers, $\tau_p$ is the cumulative probability threshold for retention, and $B$ is the block size used by block-sparse attention, which set to the number of visual tokens per frame $N_v$ (or a divisor of it) so that block boundaries align with frame boundaries.

\textbf{Evidence Decomposition.}
We average the attention maps across heads while preserving the layer dimension. 
Since the block boundaries are aligned with frame boundaries, the visual blocks corresponding to each anchor frame can be directly located in $\mathbf{A}^{(l)}$.
Then we extract three complementary evidence cues $\mathbf{A}_{qf}$, $\mathbf{A}_{ff}$ and $\mathbf{A}_{if}$ through deterministic slicing and aggregation, without recomputing attention or introducing learnable parameters.
As illustrated in Fig.~\ref{fig:main}, query-frame cue $\mathbf{A}_{qf}\in\mathbb{R}^{L\times T}$ is obtained by averaging the attention from the complete query-token group to the visual blocks of each anchor frame. Intra-frame cue $\mathbf{A}_{if} \in \mathbb{R}^{L \times T \times N_v \times N_v}$ retains the token-level causal attention within each anchor frame, retaining its spatial structure.
Cross-frame cue $\mathbf{A}_{ff}\in\mathbb{R}^{L\times (T-1)\times (T-1)}$ captures the attention from each anchor frame to its preceding anchor frames, where each pairwise score is computed by averaging the attention between their visual blocks. 
Since $\mathbf{A}_{ff}$ is indexed by frame pairs rather than individual timestamps, we aggregate it along both directions: for each anchor frame, the attention it assigns to preceding frames and the attention it receives from subsequent frames are averaged at each layer, yielding timestamp-aligned temporal evidence $\mathbf{E}_{ff}$. For the other two cues, $\mathbf{E}_{qf}$ directly inherits the timestamp-aligned query-frame relevance scores from$\mathbf{A}_{qf}$, while $\mathbf{E}_{if}$ is obtained by vectorizing the layer-wise intra-frame attentions of $\mathbf{A}_{if}$ at each timestamp.


\subsection{Distribution-Aware Dynamic Sampling}
Given the evidence priors extracted from the target MLLM, we model visual selection as a distribution-aware sampling policy over the timestamp sequence. 
Concretely, the selector takes $\mathbf{E}_{qf}$, $\mathbf{E}_{ff}$, and $\mathbf{E}_{if}$ as inputs, and predicts three timestamp-wise categorical distributions: $\pi_{\mathrm{keep}}$ over timestamp retention $z_t\in\{0,1\}$, which retains ($z_t=1$) or discards ($z_t=0$) timestamp $t$; $\pi_{\mathrm{rate}}$ over local sampling level $\rho_t\in\mathcal{R}$, which specifies the number of frames uniformly sampled from the temporal segment corresponding to timestamp $t$; and $\pi_{\mathrm{res}}$ over spatial resolution $\lambda_t\in\mathcal{L}$, which determines the resolution to which the sampled frames from this segment are resized. Here, $\mathcal{R}$ and $\mathcal{L}$ denote predefined candidate sets.
To fuse the three cues, we project each of them into a common embedding space with the timestamp dimension preserved.
The three projected features are concatenated along the feature dimension and processed by a shared lightweight MLP to obtain the fused hidden representations $\mathbf{h}$.
Three independent prediction heads then take $\mathbf{h}$ as input and output  $\pi_{\mathrm{keep}}(\cdot\mid \mathbf{h})$, $\pi_{\mathrm{rate}}(\cdot\mid \mathbf{h})$, and $\pi_{\mathrm{res}}(\cdot\mid \mathbf{h})$ for all $T$ anchor timestamps in one forward pass. The joint policy over the sampling decisions is formulated as:
\begin{equation}
\begin{aligned}
\pi_s\!\left(\{\mathbf{u}_t\}_{t=1}^T \mid Q, V_c\right)
={}&\prod_{t=1}^{T}
\pi_{\mathrm{keep}}(z_t\mid\mathbf{h})\\
&\quad\cdot
\pi_{\mathrm{rate}}(\rho_t\mid\mathbf{h})\,
\pi_{\mathrm{res}}(\lambda_t\mid\mathbf{h}).
\end{aligned}
\end{equation}

The sampled decisions are then applied to the original video $V$. For a retained anchor ($z_t=1$), we uniformly sample $\rho_t$ frames from its corresponding temporal segment and resize them to the resolution specified by $\lambda_t$; otherwise, the segment is skipped. Finally, the sampled frames are arranged chronologically to form the multi-scale frame sequence $V_s$ for downstream MLLM inference.


\subsection{Policy Optimization with GRPO}
For each training instance $(Q,V_c)$, we sample a group of $G$ candidate decision sequences and evaluate them using target MLLM. The reward of our method is composed of performance and efficiency rewards. Specifically, the performance reward $R_{\mathrm{per}}$ is binary, assigning $1$ to a correct answer and $0$ otherwise. We define the efficiency reward as a weighted combination of a retention term and a token-cost term:
\begin{equation}
R_{\mathrm{eff}}=\eta\left(1-\frac{\sum_{t=1}^{T} z_t}{T}\right) + (1-\eta)\left(1-\frac{C_{V_s}}{C_{\max}}\right),
\end{equation}
where $\eta$ balances retention and token-cost; $C_{V_s}$ is the total number of visual tokens in $V_s$, jointly determined by $\rho_t$ and $\lambda_t$ selected across all retained timestamps; and $C_{\max}$ is the corresponding upper bound obtained by applying maximum sampling level and resolution at same retained timestamps.

When all candidates in a group are incorrect, the performance rewards are identical, while the efficiency rewards may still differ. A naive additive two rewards would therefore rank the candidates solely according to their visual cost, encouraging a low-cost policy.
To avoid this, we activate the efficiency reward only if the group contains a correct candidate. Concretely, for a group of $G$ candidates, we define:
\begin{equation}
C_{\mathrm{acc}} \triangleq \mathds{1}\{\exists j \in \{1, \dots, G\} : R_{\mathrm{per},j}= 1\},
\end{equation}
where $R_{\mathrm{per},j}$ is the performance reward of the $j$-th candidate. The overall reward of the $i$-th candidate is defined as:
\begin{equation}
R_i = R_{\mathrm{per},i} + \alpha \, C_{\mathrm{acc}} \, R_{\mathrm{eff},i},
\end{equation}
where $\alpha$ controls the strength of the efficiency reward.
Finally, we optimize the selector policy using GRPO by maximizing the following objective:
\begin{equation}
\begin{aligned}
J(\theta)
= \mathbb{E}\Bigg[
\frac{1}{G}\sum_{i=1}^{G}
\min\Bigl(
r_i(\theta)A_i, 
\operatorname{clip}\bigl(
r_i(\theta),1-\epsilon,1+\epsilon
\bigr)A_i
\Bigr) 
\Bigg],
\end{aligned}
\end{equation}
where 
$ A_{i}=\frac{R_{i}-\operatorname{mean}\!\left(R_{1},R_{2},\ldots,R_{G}\right)}
{\operatorname{std}\left(R_{1},R_{2},\ldots,R_{G}\right)+\delta} $
is the group-relative advantage computed from the group rewards, $\delta$ is a small constant for numerical stability. The importance ratio is defined as $r_i(\theta)=\frac{\pi_{s,\theta}(\mathbf{u}_{1:T}^{(i)}\mid Q,V_c)}{\pi_{s,\theta_{\mathrm{old}}}(\mathbf{u}_{1:T}^{(i)}\mid Q,V_c)}$, which compares the probabilities assigned to the sampled decision sequence by the current and old selector policies. Here, $\theta_{\mathrm{old}}$ denotes the selector parameters before the current optimization update, and $\epsilon$ is the clipping threshold. Notably, we exclude the KL-divergence term used in GRPO, as our selector is trained from scratch and has no reference model to regularize against.

\begin{table*}[t]
\centering
\resizebox{\linewidth}{!}
{
\begin{tabular}{lccccccc}
    \toprule
    \multirow{2}{*}{Model}  & \multirow{2}{*}{\shortstack[c]{Avg. Selected \\ Tokens}} & \multirow{2}{*}{LLM Size}&\multirow{2}{*}{Selector} & {LongVideoBench}&{MLVU} & \multicolumn{2}{c}{Video-MME}\\
    \cmidrule(lr){5-8} 
     &  & & &Val& Dev & Long & Avg \\
    \midrule\midrule
    \multicolumn{8}{l}{\textit{Closed-Source MLLM}} \\
    \color{gray} GPT-4o~\cite{hurst2024gpt}    & \color{gray} - & - & Uniform & \color{gray} 66.7  &\color{gray}64.6& \color{gray} 65.3 & \color{gray} 71.9 \\
    \color{gray} GPT-4V~\cite{gpt4v}   & \color{gray} - & - & Uniform & \color{gray} 61.3 &\color{gray}49.2& \color{gray} 53.5 & \color{gray} 59.9 \\
    \color{gray} Gemini-1.5-Flash~\cite{team2023gemini}   & \color{gray} -& - & Uniform & \color{gray} 61.6 &- & \color{gray} 61.1 & \color{gray} 70.3 \\
    \color{gray} Gemini-1.5-Pro~\cite{team2023gemini}   & \color{gray} - & -& Uniform & \color{gray} 64 &- & \color{gray} 67.4 & \color{gray} 75.0 \\
    \midrule
    \multicolumn{8}{l}{\textit{Open-Source MLLM}} \\
    Video-LLaVA~\cite{lin2024video}& 2048 &7B&Uniform&-&36.2&-&39.9\\
    Qwen-VL~\cite{bai2023qwen} & 2048 & 7B  & Uniform & - &- & 37.8 & 41.1\\
    Oryx-1.5~\cite{liu2024oryx}  & 14400 & 7B  & Uniform & 56.3 &- & 51.2 & 58.8 \\
    LLaVA-Onevision~\cite{li2024llava}   & 6272 & 7B  & Uniform & 56.4&64.7 & 46.7 & 58.2\\
    NVILA~\cite{liu2025nvila}   & 8192  & 7B& Uniform &57.7&70.1&54.8&64.2\\
    Apollo~\cite{zohar2025apollo} & 2FPS & 7B  & Uniform & 58.5 &68.7& -  & 61.3\\
    LongVU~\cite{shen2024longvu}  & 1FPS& 7B  & DINOv2 &-&65.4&-&60.6\\


    \midrule
    LLaVA-Video-7B*   & 3360 & 7B  & Uniform & 57.4  & 64.4 & 51.3 & 60.3 \\
    +AKS~\cite{tang2025adaptive}  & 3360 & 7B & CLIP &  61.6 & - & - & 62.2  \\
    +TSPO~\cite{tang2025tspo}*  & 3360 & 7B & TSPO-0.4B &  61.4 & 68.2 & 52.9 & 62.1  \\
    \rowcolor{resultcolor}
    +EviSelect  & \textbf{1701}& 7B &\textbf{0.1B}&  \textbf{62.1} & \textbf{69.1} & \textbf{53.7} & \textbf{64.0} \\

    \midrule
    Qwen2.5-VL-7B* & 2912 & 7B & Uniform & 55.4 & 54.3 & 48.9  & 57.1 \\
    +FastV + DyToK~\cite{li2025less}& 1568 & 7B & — & 54.5 & 43.3 & - & 58.8 \\
    +Q-Frame~\cite{zhang2025q}  & 2912 & 7B & CLIP & 57.37 & 56.81&49.02 & -  \\
    +DIG~\cite{li2025divide}  & 2912 & 7B & DINOv2 & 57.89 & 63.98&51.93 & -  \\
    +TSPO~\cite{tang2025tspo}*  & 2912 & 7B & TSPO-0.4B & 58.1 & 65.1&52.1 & 59.1  \\
    \rowcolor{resultcolor}
    +EviSelect  & \textbf{1439} & 7B & \textbf{0.1B} & \textbf{59.2} & \textbf{66.1} & \textbf{53.0} & \textbf{60.8}  \\
    
    \bottomrule
    \end{tabular}
}
\caption{Evaluation of long-form video understanding on three commonly used benchmarks. ``*'' denotes our reproduced results.}
\label{tab:main_result}
\end{table*}

\section{Experiments}
\label{sec:experiment}

\subsection{Benchmarks}
To evaluate the effectiveness of our method, we conduct extensive experiments on three widely used long-form video understanding benchmarks: (1) MLVU~\cite{zhou2024mlvu}, a multi-task benchmark that ranges from 3 minutes to 2 hours (average video duration: 12 minutes) and covers diverse video genres such as movies and surveillance footage. We evaluate on the ``M-Avg'' portion of the ``Dev'' split, following~\cite{zhang2024llava}. (2) LongVideoBench~\cite{wu2024longvideobench}, evaluated on the validation set without subtitles (average video duration: 12 minutes), containing 1,337 QA pairs across 17 fine-grained categories. (3) Video-MME~\cite{fu2025video} comprises 900 videos with durations split into short ($<$ 2min), medium (4$\sim$15min), and long (30$\sim$60min), totaling 2,700 QA pairs (average video duration: 17 minutes).

\subsection{Implementation Detail}
We evaluate our framework on two MLLMs: Qwen2.5-VL-7B~\cite{bai2025qwen2} and LLaVA-Video-7B~\cite{zhang2024video}. All experiments are conducted on 4 NVIDIA H100 GPUs. Following TSPO~\cite{tang2025tspo}, our lightweight selector (0.1B parameters) is trained on the same 10K subset of LLaVA-Video-178K dataset~\cite{zhang2024video} for one epoch, with a learning rate of 5$\times 10^{-4}$ and a batch size of 1. In the GRPO algorithm, we set $G=8$, the efficiency reward coefficient $\alpha=0.1$, the retention–budget trade-off $\eta=0.5$, and the clipping threshold to $\epsilon=0.2$. 
During evidence grounding, we uniformly divide each video into consecutive 2-second temporal segments and use the central timestamp of each segment as its anchor. Each anchor frame is spatially downsampled and represented by $N_v=20$ visual tokens.
The top-$p$ threshold is set to $\tau_p = 0.97$ and the block size is fixed by default at $B=20$. To accommodate the intrinsic temporal patch size of the visual encoder (e.g., a temporal patch size of 2 in Qwen2.5-VL~\cite{bai2025qwen2}), we define the candidate set for sampling rate as $\mathcal{R}=\{1,2,4,8\}$. The spatial resolution candidate set $\mathcal{L}$ comprises 4 levels $90\times160$, $360\times640$, $540\times960$, and $720\times1280$.

\subsection{Main Results}

\textbf{Comparison with Existing Methods.}
As shown in Tab.\ref{tab:main_result}, LLaVA-Video-7B equipped with EviSelect achieves state-of-the-art performance across three general long video benchmarks, while using fewer selected visual tokens. Compared with TSPO~\cite{tang2025tspo}, our EviSelect achieves performance improvements of 1.9\% on Video-MME~\cite{fu2025video}, 0.9\% on MLVU~\cite{zhou2024mlvu}, and 0.7\% on LongVideoBench~\cite{wu2024longvideobench}, while using only 1701 tokens on average, reducing the answer-stage visual-token budget by 49\% compared with TSPO (3360 tokens). Moreover, EviSelect using Qwen2.5-VL-7B~\cite{bai2025qwen2} also brings consistent improvements over TSPO, demonstrating strong generalization across different MLLM. Meanwhile, it also achieves competitive performance in challenging ``Long" subset of Video-MME. This indicates its robust ability to localize query-relevant content even under extremely long contexts, which benefits from our use of the evidence priors of the target MLLM to dynamically identify key timestamps.

\begin{table}[t]
\centering
\resizebox{\columnwidth}{!}{%
\begin{tabular}{lcccc}
\toprule
\multicolumn{1}{c}{\multirow{2}{*}[-0.5ex]{Method}}
& \multirow{2}{*}[-0.5ex]{\shortstack[c]{Selected \\ Tokens}}
& \multicolumn{3}{c}{Latency (s)} \\
\cmidrule(lr){3-5}
& & Selection & MLLM & Overall \\
\midrule
Uniform   & 3360  & 0    & 0.78 & \textbf{0.78} \\
CoS~\cite{hu2025cos}       & 13440 & 28.4 & 2.70 & 31.10 \\
TSPO~\cite{tang2025tspo}      & 3360  & 9.2  & 0.78 & 9.98 \\
\rowcolor{resultcolor}
EviSelect & \textbf{1701} & \textbf{2.1}
          & \textbf{0.45} & {2.55} \\
\bottomrule
\end{tabular}
}
\caption{End-to-end inference efficiency.}
\label{tab:inference_time}
\end{table}
\noindent\textbf{Inference Efficiency.}
As shown in Tab.~\ref{tab:inference_time}, \emph{Selected tokens} denotes the average number of visual tokens selected for and passed to the target MLLM. For EviSelect, \emph{Selection} includes the complete evidence-grounding pass, selector inference, frame resampling, and final input construction. \emph{MLLM} measures the subsequent time for the target MLLM to process the input and generate the answer, and \emph{Overall} reports the end-to-end latency as the sum of these two stages.
Compared with TSPO~\cite{tang2025tspo}, EviSelect reduces the selection time from 9.2\,s to 2.1\,s and the answer-stage visual-token cost from 3360 to 1701. It also reduces MLLM inference time from 0.78\,s to 0.45\,s, resulting in an end-to-end latency of 2.55\,s and a $3.9\times$ speedup.

\subsection{Ablation Study}

\begin{table}[t]
\centering
    \setlength{\tabcolsep}{4pt} 
    \resizebox{0.98\linewidth}{!}{
    \begin{tabular}{cccccc}
    \toprule
    Retention & Adaptive Frame & Adaptive Spatial & \multirow{2}{*}{Token} & \multirow{2}{*}{LVB} & \multirow{2}{*}{MLVU} \\
    Strategy & Sampling & Resolution &&&\\
    \midrule\midrule
    \multicolumn{1}{c}{\textcolor[gray]{0.8}{\ding{55}}} & \multicolumn{1}{c}{\textcolor[gray]{0.8}{\ding{55}}} & \multicolumn{1}{c}{\textcolor[gray]{0.8}{\ding{55}}} & 3360 & 57.4 & 64.4 \\
    \multicolumn{1}{c}{\ding{51}} & \multicolumn{1}{c}{\textcolor[gray]{0.8}{\ding{55}}} & \multicolumn{1}{c}{\textcolor[gray]{0.8}{\ding{55}}} & 2100 & 58.5 & 65.4  \\
    \multicolumn{1}{c}{\ding{51}} & \multicolumn{1}{c}{\ding{51}} & \multicolumn{1}{c}{\textcolor[gray]{0.8}{\ding{55}}} & 2730 & 59.1 & 67.2 \\
    \multicolumn{1}{c}{\ding{51}} & \multicolumn{1}{c}{\textcolor[gray]{0.8}{\ding{55}}} & \multicolumn{1}{c}{\ding{51}} & 2048 & 60.3 & 67.9 \\
    
    \rowcolor{resultcolor}
    \multicolumn{1}{c}{\ding{51}} & \multicolumn{1}{c}{\ding{51}} & \multicolumn{1}{c}{\ding{51}} & \textbf{1701} & \textbf{62.1} & \textbf{69.1} \\

    \bottomrule
    \end{tabular}}
    \caption{Ablation study of different components. }
\label{tab:component}
\end{table}

\noindent\textbf{Effect of Different Components.}
To evaluate the contribution of each component in EviSelect, we conduct comprehensive ablation studies on MLVU \cite{zhou2024mlvu} and LongVideoBench~\cite{wu2024longvideobench} using LLaVA-Video-7B~\cite{zhang2024video} as the MLLM. As shown in Tab.~\ref{tab:component}, the retention strategy alone improves accuracy over the uniform-sampling (64.4\%$\rightarrow$65.4\% on MLVU and 57.4\%$\rightarrow$58.5\% on LongVideoBench). This indicates that evidence-driven timestamp retention is crucial for discarding redundant segments while preserving critical evidence, whereas uniform sampling frequently misses query-relevant visual cues. 

Moreover, based on timestamp retention, both adaptive strategies bring further gains: adaptive frame sampling improves accuracy from 65.4\% to 67.2\%, and adaptive spatial resolution improves it to 67.9\%, indicating that a fixed token allocation fails to accommodate the uneven distribution of temporal and spatial information. 
Notably, adaptive spatial resolution achieves higher accuracy with fewer tokens than adaptive frame sampling. We attribute this to the higher redundancy in spatial dimension: the semantics of most frames are well preserved at low resolution, so adaptive spatial resolution removes redundant tokens with little semantic loss, whereas densifying frames incurs a larger token overhead for a smaller gain. Combining all components, which reduce redundancy along orthogonal dimensions, achieves the best accuracy (69.1\% on MLVU and 62.1\% on LongVideoBench) at the lowest cost of 1701 tokens.

\begin{figure}[!t]
\centering
\includegraphics[width=0.6\linewidth]{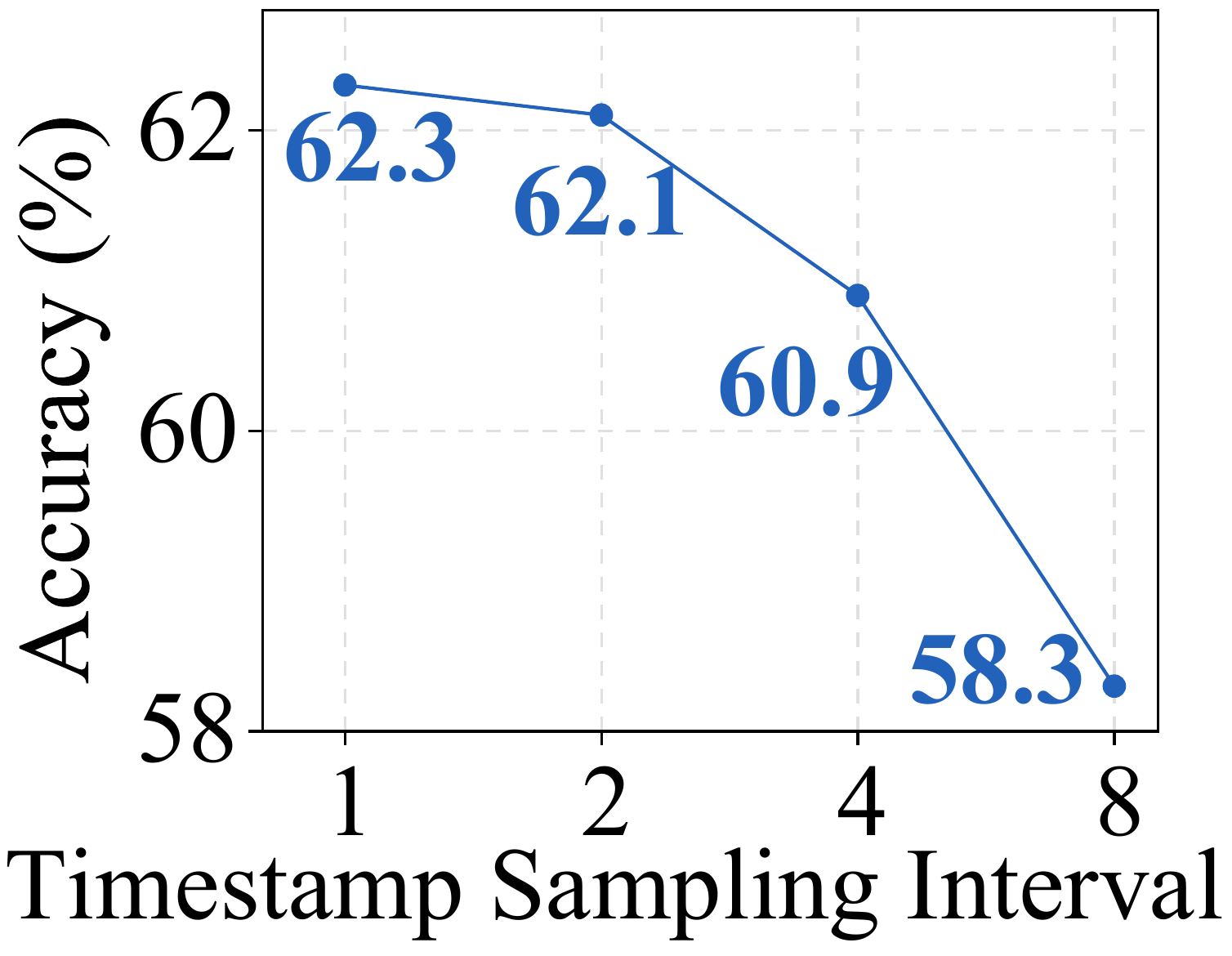}
\caption{Effect of the anchor timestamp sampling interval.}
\label{fig:timestamp}
\end{figure}

\begin{table*}[t!]
    \centering
    \begin{minipage}[t]{0.5\textwidth}
    \centering
    \resizebox{0.8\linewidth}{!}{%
        \begin{tabular}{*{4}{wc{1.7em}}cc}
            \toprule
            \multicolumn{4}{c}{Local Sampling Rate ($\rho_t$)}
            & \multirow{2}{*}{Token}
            & \multirow{2}{*}{Acc} \\
            8 & 4 & 2 & 1 & & \\
            \midrule\midrule
            \ding{51}
            & \textcolor[gray]{0.8}{\ding{55}}
            & \textcolor[gray]{0.8}{\ding{55}}
            & \textcolor[gray]{0.8}{\ding{55}}
            & 3045 & 60.7 \\

            \textcolor[gray]{0.8}{\ding{55}}
            & \ding{51}
            & \textcolor[gray]{0.8}{\ding{55}}
            & \textcolor[gray]{0.8}{\ding{55}}
            & 2048 & 60.3 \\

            \textcolor[gray]{0.8}{\ding{55}}
            & \textcolor[gray]{0.8}{\ding{55}}
            & \ding{51}
            & \textcolor[gray]{0.8}{\ding{55}}
            & 1624 & 58.5 \\

            \textcolor[gray]{0.8}{\ding{55}}
            & \textcolor[gray]{0.8}{\ding{55}}
            & \textcolor[gray]{0.8}{\ding{55}}
            & \ding{51}
            & 1356 & 55.2 \\

            \ding{51}
            & \ding{51}
            & \textcolor[gray]{0.8}{\ding{55}}
            & \textcolor[gray]{0.8}{\ding{55}}
            & 2458 & 61.0 \\

            \ding{51}
            & \ding{51}
            & \ding{51}
            & \textcolor[gray]{0.8}{\ding{55}}
            & 1950 & 61.5 \\

            \rowcolor{resultcolor}
            \ding{51} & \ding{51} & \ding{51} & \ding{51}
            & \textbf{1701} & \textbf{62.1} \\
            \bottomrule
        \end{tabular}%
    }
    \captionof{table}{Ablation on the frame-sampling candidate set $\mathcal{R}$.}
    \label{tab:multi-rate}
    \end{minipage}\hfill%
    \begin{minipage}[t]{0.5\textwidth}
    \centering
    \resizebox{0.8\linewidth}{!}{%
        \begin{tabular}{cccccc}
            \toprule
            \multicolumn{4}{c}{Frame Resolution ($\lambda_t$)}
            & \multirow{2}{*}{Token}
            & \multirow{2}{*}{Acc} \\
            720P & 540P & 360P & 90P & & \\
            \midrule\midrule
            \ding{51}
            & \textcolor[gray]{0.8}{\ding{55}}
            & \textcolor[gray]{0.8}{\ding{55}}
            & \textcolor[gray]{0.8}{\ding{55}}
            & 4261 & 59.3 \\

            \textcolor[gray]{0.8}{\ding{55}}
            & \ding{51}
            & \textcolor[gray]{0.8}{\ding{55}}
            & \textcolor[gray]{0.8}{\ding{55}}
            & 2735 & 59.1 \\

            \textcolor[gray]{0.8}{\ding{55}}
            & \textcolor[gray]{0.8}{\ding{55}}
            & \ding{51}
            & \textcolor[gray]{0.8}{\ding{55}}
            & 2059 & 57.5 \\

            \textcolor[gray]{0.8}{\ding{55}}
            & \textcolor[gray]{0.8}{\ding{55}}
            & \textcolor[gray]{0.8}{\ding{55}}
            & \ding{51}
            & 1563 & 54.9 \\

            \ding{51}
            & \ding{51}
            & \textcolor[gray]{0.8}{\ding{55}}
            & \textcolor[gray]{0.8}{\ding{55}}
            & 3218 & 60.2 \\

            \ding{51}
            & \ding{51}
            & \ding{51}
            & \textcolor[gray]{0.8}{\ding{55}}
            & 2879 & 61.1 \\

            \rowcolor{resultcolor}
            \ding{51} & \ding{51} & \ding{51} & \ding{51}
            & \textbf{1701} & \textbf{62.1} \\
            \bottomrule
        \end{tabular}%
    }
    \captionof{table}{Ablation on the spatial-resolution candidate set $\mathcal{L}$.}
    \label{tab:multi-res}
    \end{minipage}

    \par\vspace{1em}
    \resizebox{0.8\linewidth}{!}{%
        \begin{tabular}{lcccccccc}
            \toprule
            \multirow{2}{*}{Model}
            & \multirow{2}{*}{\shortstack{Selected \\ tokens}}
            & \multirow{2}{*}{\shortstack{LLM \\ Size}}
            & \multirow{2}{*}{Selector}
            & {LongVideoBench} & {MLVU} & \multicolumn{2}{c}{Video-MME} \\
            \cmidrule(lr){5-8}
            & & & & Val & Dev & Long & Avg \\
            \midrule\midrule
            \midrule
            LLaVA-Video-7B* & 3360 & 7B & Uniform & 57.4 & 64.4 & 51.3 & 60.3 \\
            Qwen$\rightarrow$ {LLaVA} & {1659} & 7B & {0.1B} & {59.8} & {67.5} & {52.5} & {62.1} \\
            \rowcolor{resultcolor}
            EviSelect & \textbf{1701} & 7B & \textbf{0.1B} & \textbf{62.1} & \textbf{69.1} & \textbf{53.7} & \textbf{64.0} \\

            \mydashline{0mm}{121.5mm}
            Qwen2.5-VL-7B* & 2912 & 7B & Uniform & 55.4 & 54.3 & 48.9 & 57.1 \\
            LLaVA $\rightarrow$ {Qwen} & {1475} & 7B & {0.1B} & {57.5} & {62.2} & {51.6} & {59.7} \\
            \rowcolor{resultcolor}
            EviSelect & \textbf{1439} & 7B & \textbf{0.1B} & \textbf{59.2} & \textbf{66.1} & \textbf{53.0} & \textbf{60.8} \\
            \bottomrule
        \end{tabular}%
    }
    \captionof{table}{Cross-backbone transfer of the selector.}
    \label{tab:trans}
\end{table*}

\noindent\textbf{Effect of Anchor Interval.}
Fig.~\ref{fig:timestamp} examines the effect of the anchor interval used in evidence grounding on LongVideoBench~\cite{wu2024longvideobench}. The default 2-second interval achieves 62.1\% accuracy, closely matching the denser 1-second setting (62.3\%), while requiring only about half the probing-token cost, as the number of anchor frames doubles when the interval is halved. In contrast, coarser intervals degrade accuracy noticeably, dropping to 60.9\% at 4 seconds and 58.3\% at 8 seconds, since the sparser anchor sequence fails to cover short events and abrupt state changes, leaving the selector without sufficient evidence for precise temporal localization. We therefore adopt the 2-second interval by default, as it strikes the best accuracy--cost balance.

\noindent\textbf{Effect of Multi-level Temporal Sampling.}
Tab.~\ref{tab:multi-rate} compares different configurations of $\mathcal{R}$ used by adaptive frame sampling on  LongVideoBench~\cite{wu2024longvideobench}. Enabling all four candidates yields the best accuracy at a lower token cost: compared with the best single-candidate configuration, it improves accuracy by 1.4\% while reducing the token cost from 3045 to 1701. Accuracy also improves consistently as more candidates become available. These results indicate that no fixed sampling rate suits both fast actions and slowly changing scenes, whereas adaptive frame sampling adapts to the varying temporal dynamics across video segments.

\noindent\textbf{Effect of Multi-level Spatial Resolution.}
Tab.~\ref{tab:multi-res} examines different configurations of the candidate set $\mathcal{L}$ used by adaptive spatial resolution on  LongVideoBench~\cite{wu2024longvideobench}. Enabling all resolution candidates achieves the highest accuracy (62.1\%) at a lower token cost. In contrast, using only the lower resolution ($90\times160$) is cheap but drops accuracy to 54.9\%, as low-resolution frames lose subtle visual evidence; using only the highest resolution ($720\times1280$) reaches merely 59.3\% while incurring the largest token cost (4261), since it spends many tokens on background regions that provide little information and may distract the model. These results suggest that effective resolution allocation is not about uniformly increasing image quality, but about assigning resolution adaptively on demand.

\noindent\textbf{Cross-backbone Transfer of Evidence-Grounded Selector.}
As shown in Tab.~\ref{tab:trans}, we evaluate cross-backbone transfer by swapping the trained selectors between LLaVA-Video-7B and Qwen2.5-VL-7B without further training. The transferred selectors consistently outperform uniform sampling while using fewer answer-stage tokens, improving Video-MME Avg from 60.3\% to 62.1\% for LLaVA and from 57.1\% to 59.7\% for Qwen. However, they still underperform native EviSelect results, indicating that the learned selection policy is partially transferable but still backbone-dependent.


\begin{table}[t]
    \centering
    
    \setlength{\tabcolsep}{5pt}
    \resizebox{1\linewidth}{!}{%
    \begin{tabular}{lcccccc}
        \toprule
        Cue & Pear.$\uparrow$ & Spear.$\uparrow$ & JS$\downarrow$
        & Top-10\%$\uparrow$ & Keep & Mass-rec.$\uparrow$ \\
        \midrule
        $\mathbf{A}_{qf}$ & 0.89 & 0.86 & 0.022 & 0.75 & 0.60 & 0.985 \\
        $\mathbf{A}_{ff}$ & 0.89 & 0.87 & 0.015 & 0.76 & 0.52 & 0.987 \\
        $\mathbf{A}_{if}$ & 0.87 & 0.83 & 0.002 & 0.62 & -- & -- \\
        \midrule
        Random & 0.00 & 0.00 & $\sim$0.10 & 0.10 & -- & -- \\
        \bottomrule
    \end{tabular}%
    }
\caption{Alignment between sparse and dense attention maps.}
\label{tab:sparse_and_dense}
\end{table}

\noindent\textbf{Reliability of Sparse Attention Maps.}
To verify whether attention maps obtained from sparse prefilling reliably approximate their dense counterparts, we compare the two on 200 sampled videos. As shown in Tab.~\ref{tab:sparse_and_dense}, all three sparse cues achieve Pearson correlations of 0.87--0.89 with dense attention, substantially exceeding the random reference. Moreover, $\mathbf{A}_{qf}$ and $\mathbf{A}_{ff}$ recover over 98\% of the dense-attention mass while retaining only about half of the positions, confirming that sparse probing captures the high-mass evidence rather than depending on near-complete coverage.

\begin{table}[t]
    \centering
    
    \setlength{\tabcolsep}{5pt}
    \resizebox{1\linewidth}{!}{%
    \begin{tabular}{lccc}
        \toprule
        Evidence & LVB$\uparrow$ & MLVU$\uparrow$ & Video-MME$\uparrow$ \\
        \midrule
        CLIP~\cite{radford2021learning} & 60.3 & 66.7 & 61.9 \\
        Random prior & 56.4 & 59.8 & 58.2 \\
        \rowcolor{resultcolor}
        Internal sparse-prefill & \textbf{62.1} & \textbf{69.1} & \textbf{64.0} \\
        \bottomrule
    \end{tabular}%
    }
\caption{Evidence-source ablation.}
\label{tab:clip_ablation}
\end{table}

\noindent\textbf{Effect of the Evidence Source.}
To study the effect of the evidence source, we independently train three selectors from scratch with internal, CLIP, and random priors under otherwise identical settings.
As shown in Tab.~\ref{tab:clip_ablation}, the internal sparse-prefill prior achieves 65.1 average accuracy, outperforming the CLIP prior by 2.1\%, while random prior degrades the average to 58.1. These results indicate that the selector benefits directly from the quality of its input evidence, and that evidence from the target MLLM itself provides more faithful guidance than from an external retriever.

\section{Conclusion}
In this paper, we present \textbf{EviSelect}, a framework that leverages the target MLLM's internal attention priors for distribution-aware spatiotemporal sampling. Through sparse prefilling on compressed inputs, EviSelect efficiently approximates attention evidence, which guides a lightweight selector to predict retention, local sampling rate, and spatial resolution policies. Optimized end-to-end via GRPO under a joint accuracy–efficiency reward, this probabilistic policy dynamically aligns visual inputs with the target MLLM's intrinsic requirements. Experiments on three long-video benchmarks show that EviSelect consistently outperforms prior methods while using about 50\% fewer answer-stage visual tokens on average and achieving a \(3.9\times\) end-to-end speedup.

\textbf{Limitations.}
Since evidence priors are extracted from the internal attention of the target MLLM, our EviSelect may have limited generalization when transferred to other MLLMs. Moreover, it is not applicable to closed-source MLLMs where internal attentions are inaccessible.

\bibliography{aaai2027}

\clearpage
\appendix
\twocolumn[
  \vbox{%
    \hsize\textwidth
    \centering
    {\LARGE\bfseries Supplementary Materials\par}%
    \vskip 0.2in
  }%
]

\section{Implementation Details}
\label{sec_supp:implement}

\subsection{Backbone-Specific Resolution Policy.}
We use a unified spatial policy across backbones, while its concrete implementation depends on the visual encoder. For Qwen2.5-VL, the four spatial levels are implemented by resizing the input frames to $720{\times}1280$, $540{\times}960$, $360{\times}640$, and $90{\times}160$, respectively. In contrast, LLaVA-Video maps every input frame to a fixed number of visual tokens, so changing the raw input resolution does not produce different visual-token budgets. We therefore implement the same four levels after visual encoding using bilinear-interpolation pooling with strides of 1, 2, 4, and 8, respectively. For LLaVA-Video, the resolution labels used in the main paper refer to these equivalent post-encoding visual-token-budget levels rather than input resolutions.

\subsection{Additional Training Details.}
To stabilize early training under sparse correctness signals, we adopt a staged reward schedule. The efficiency-reward coefficient $\alpha=0.1$ reported in the main paper denotes its target value after warm-up. During the first 15\% of the training steps, we optimize the selector using only the performance reward, corresponding to $\alpha=0$. From 15\% to 30\% of training, we linearly increase $\alpha$ from 0 to 0.1. We use the full reward with $\alpha=0.1$ for the remaining training steps. The total training cost is 133 GPU-hours on NVIDIA H100 GPUs.

During policy sampling, if all sampled retention decisions are zero, we discard the retention sample and resample the retention decisions until at least one timestamp is retained. The accepted decision sequence therefore always contains at least one retained timestamp, ensuring a non-empty visual input and a well-defined $C_{\max}$.

\subsection{Detailed Sparse Prefilling.}
We use $Q$ to denote the textual query and $\mathbf{q}_i$ to denote the projected attention query at token position $i$. Layer and head indices are omitted for clarity.

The prefilling context is organized as $X=[X_{\mathrm{sys}};X_{\mathrm{vis}};X_Q]$, consisting of a system-prompt prefix, visual tokens, and the tokenized query, respectively. The system-prompt prefix is always visible to subsequent tokens and is excluded from sparse block selection.

We partition the visual sequence in a frame-aligned manner. Each anchor frame contains $N_v$ consecutive visual tokens, and the block size $B$ is chosen such that $B$ divides $N_v$. Consequently, each frame is partitioned into $N_v/B$ contiguous blocks, and no visual block crosses a frame boundary. The query tokens are partitioned independently into consecutive blocks of size $B$, while its final block may contain fewer than $B$ tokens.

Let $K_j$ denote the set of projected key vectors associated
with the tokens in the $j$-th block. We represent the block by the mean of its keys:
\begin{equation}
\bar{\mathbf{k}}_j =
\frac{1}{|K_j|}
\sum_{\mathbf{k}_m\in K_j}\mathbf{k}_m.
\end{equation}

For each projected attention-query vector $\mathbf{q}_i$, let $\mathcal{H}_i$ denote the blocks that precede position $i$. A block belongs to $\mathcal{H}_i$ only if it ends before token position $i$ and belongs to the same input sample. We compute the block-level affinity as
\begin{equation}
g_{i,j} =
\frac{\mathbf{q}_i\bar{\mathbf{k}}_j^\top}{\sqrt{d}},
\qquad
K_j\in\mathcal{H}_i,
\end{equation}
where $d$ is the attention-head dimension. The affinities are normalized over the valid historical blocks:
\begin{equation}
p_{i,j} =
\frac{\exp(g_{i,j})}
{\sum_{K_r\in\mathcal{H}_i}\exp(g_{i,r})}.
\end{equation}

We rank the $\mathcal{H}_i$ in descending order of $p_{i,j}$ and retain the smallest subset $\mathcal{S}_i\subseteq\mathcal{H}_i$ whose cumulative probability reaches $\tau_p$, while masking out all remaining historical blocks.

\begin{figure*}[t]
\centering
\includegraphics[width=0.8\linewidth]{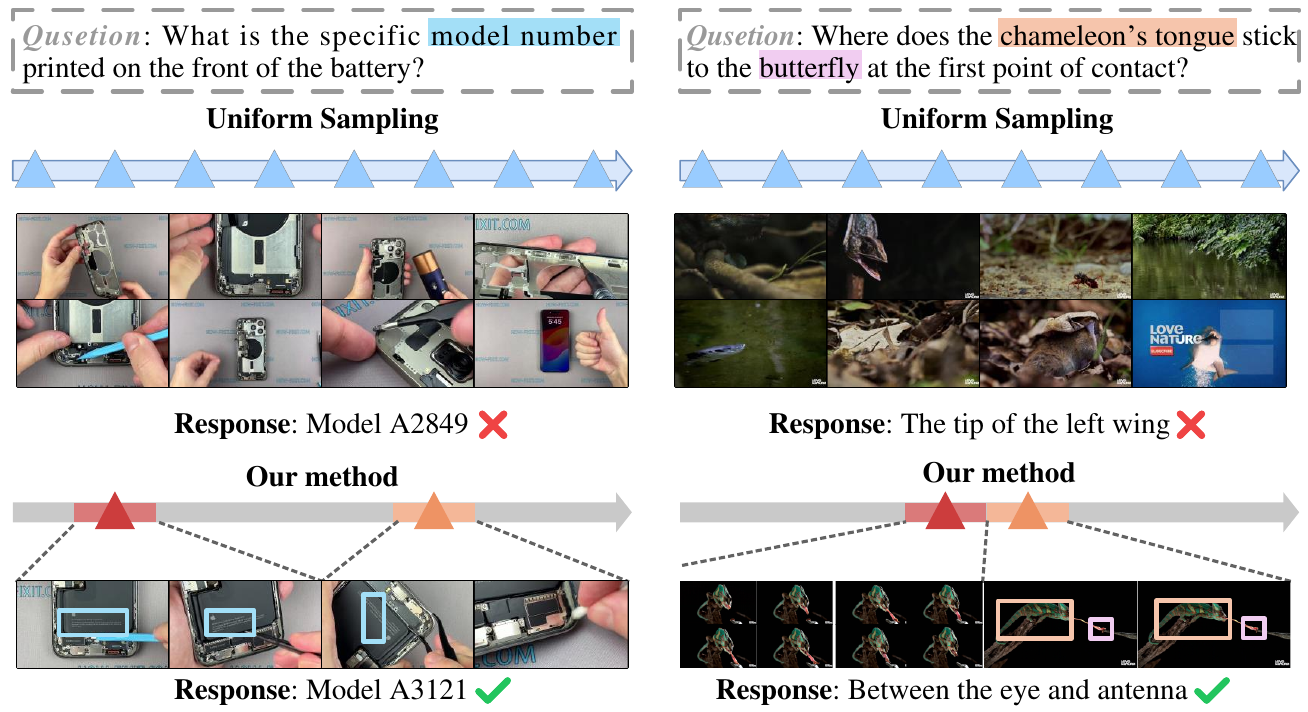} 
\caption{Qualitative comparison between uniform sampling and our method.} 
\label{fig:Visualization}
\end{figure*}

\begin{figure*}[t]
\centering
\includegraphics[width=0.8\linewidth]{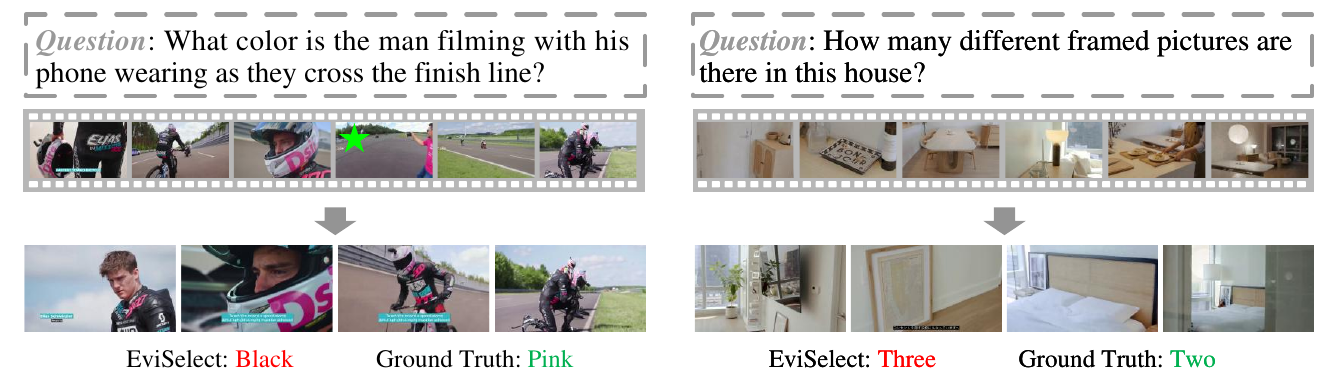}
\caption{Representative failure cases of EviSelect.}
\label{fig:supvis}
\end{figure*}

For evidence extraction, we reconstruct a coarse token-level attention map from the block-level affinities. Tokens in a retained historical block share its corresponding block-level affinity, whereas tokens in the system-prompt prefix and the causal prefix of the current block retain their exact attention logits:
\begin{equation}
a_{i,m} =
\begin{cases}
\displaystyle
\frac{\mathbf{q}_i\mathbf{k}_m^\top}{\sqrt{d}},
& m\in\mathcal{E}_i, \\[6pt]
g_{i,j},
& \mathbf{k}_m\in K_j,\;
  K_j\in\mathcal{S}_i, \\[2pt]
-\infty,
& \text{otherwise},
\end{cases}
\end{equation}
where $\mathcal{E}_i$ contains the causally valid system-prefix tokens and the causal prefix of the current block. The coarse attention map is obtained by row-wise normalization:
\begin{equation}
\widetilde{A}_{i,m} =
\frac{\exp(a_{i,m})}
{\sum_n\exp(a_{i,n})}.
\end{equation}

Applying this reconstruction at each layer and averaging over attention heads produces the layer-wise coarse attention maps $\{A^{(l)}\}_{l=1}^{L}$. Using the boundaries of the system, visual, and query-token regions, we group the frame-aligned visual blocks according to their corresponding anchor frames and extract $\mathbf{A}_{qf}$, $\mathbf{A}_{ff}$, and $\mathbf{A}_{if}$.

\section{Additional Results and Analysis}
\label{sec_supp:additional_results_analysis}

\subsection{Qualitative Results.}
Fig.~\ref{fig:Visualization} compares the keyframe selection patterns and model responses of uniform sampling and EviSelect. In the left example, when recognizing tiny text such as a battery model number, uniform sampling is query-agnostic, selecting frames from various irrelevant scenes. In contrast, our method not only localizes the relevant temporal segment but also adaptively assigns a higher resolution to the selected frames to preserve fine-grained details, achieving the correct answer with fewer frames. In the right example, the query relies on the critical moment when the chameleon shoots its tongue and catches the butterfly. Our method adaptively densifies sampling around this action while using a lower resolution to control cost, thereby capturing the relevant visual evidence with a lower overhead. These examples qualitatively illustrate that our method leverages the MLLM’s internal evidence priors to localize query-relevant timestamps, adaptively increases resolution to recognize subtle details, and adjusts the sampling rate to capture rapid temporal changes.

\subsection{Failure Case Analysis.}
As shown in Fig.~\ref{fig:supvis}, we present representative failure cases of EviSelect. In the left example, the query depends on the brief moment when the man crosses the finish line. During evidence grounding, the uniformly spaced anchor frames fail to capture this brief event. Although the MLLM can infer relatively smooth temporal changes from discrete frames, it cannot recover visual evidence that is entirely absent from the sampled input. A potential improvement is to perform an initial higher-rate or motion-aware scan, followed by adaptive refinement around high-change or high-uncertainty intervals.

In the right example, the selector retrieves frames containing all target instances, but the MLLM still fails on the counting task. The error arises from inadequate cross-frame deduplication and instance aggregation rather than missing visual evidence. Because one instance can appear across multiple frames with substantial viewpoint changes, the MLLM may count redundant observations as distinct entities. A promising direction is to incorporate cross-frame instance association and redundancy-aware selection constraints: for counting queries, the selector should retain a canonical frame for each unique instance, whereas reasoning-intensive queries may require complementary multi-view evidence.

\subsection{Hyperparameter Analysis.}
We analyze the effects of the per-frame token count, block size, and top-$p$ threshold.

\begin{figure*}[t]
\centering
\includegraphics[width=\linewidth]{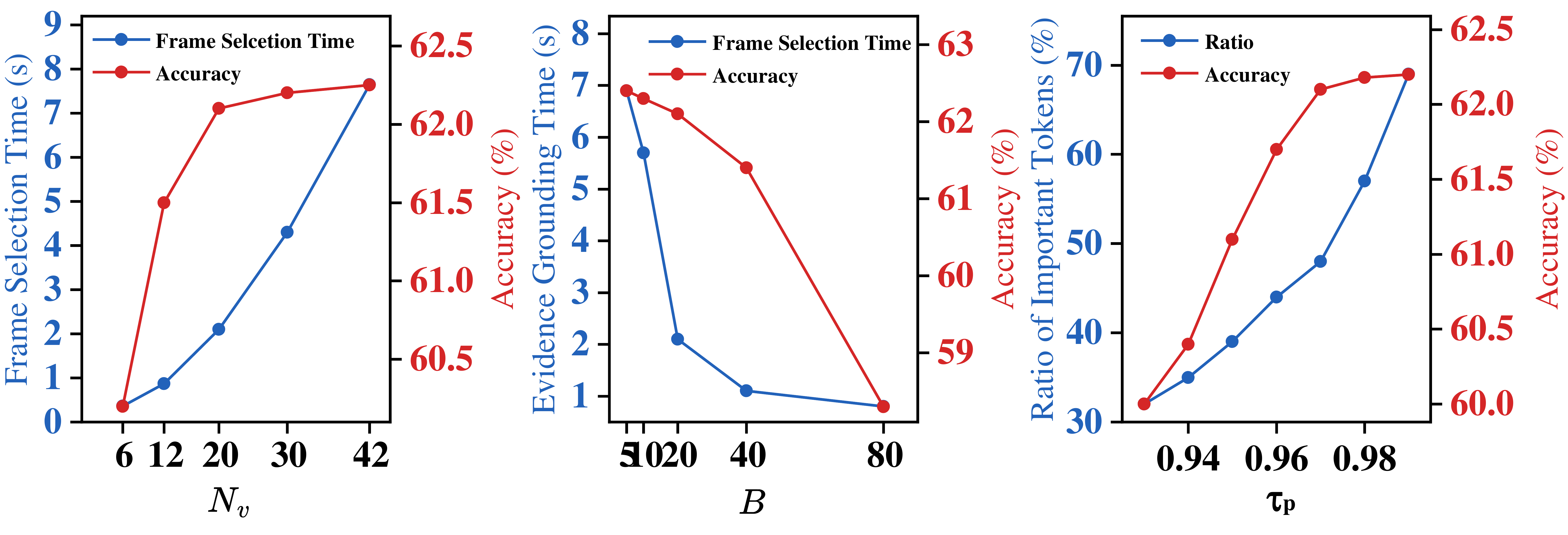}
\caption{Influence of the per-frame token count $N_v$, block size $B$, and top-$p$ threshold $\tau_p$ on LongVideoBench.}
\label{fig:discuss}
\end{figure*}

\subsection{Per-Frame Token Count $N_v$.}
The number of tokens per frame directly determines the cost of evidence grounding. As shown in the left panel of Fig.~\ref{fig:discuss}, increasing $N_v$ improves accuracy but also increases frame-selection time. The gains quickly saturate, while overly aggressive compression causes a substantial performance drop. We therefore use $N_v=20$ as a balance between accuracy and efficiency. This result suggests that the evidence-grounding stage can reliably identify relevant content from compact per-frame representations without requiring a high-fidelity reconstruction of every frame.

\subsection{Block Size $B$.}
We vary the block size over $B \in \{5,10,20,40,80\}$, with each frame represented by 20 visual tokens. The settings $B=40$ and $B=80$ intentionally allow blocks to span adjacent frames as non-frame-aligned sensitivity settings. As shown in the middle panel of Fig.~\ref{fig:discuss}, smaller blocks provide finer-grained evidence and generally improve performance, whereas larger blocks can mix tokens from different frames and blur temporal boundaries. We use $B=20$ because reducing it further substantially increases the frame-selection time while providing only marginal accuracy gains.

\subsection{Top-$p$ Threshold $\tau_p$.}
The right panel of Fig.~\ref{fig:discuss} shows how $\tau_p$ affects performance and the retained-token ratio. A lower threshold reduces computation but may discard context required for a reliable evidence prior. Conversely, performance gains quickly saturate as the threshold increases. We therefore use $\tau_p=0.97$, which provides a favorable accuracy--efficiency balance.

\end{document}